\documentclass{Interspeech2024}

\interspeechcameraready 

\title{An End-to-End Speech Summarization Using Large Language Model}

\name{Hengchao}{Shang$^*$}
\name{Zongyao}{Li$^*$}
\name{Jiaxin}{Guo}
\name{Shaojun}{Li}
\name{Zhiqiang}{Rao}
\name{\\Yuanchang}{Luo}
\name{Daimeng}{Wei}
\name{Hao}{Yang}

\address{
  $^1$Huawei Translation Service Center, Beijing, China}
\email{\{shanghengchao, lizongyao, guojiaxin1, lishaojun18,raozhiqiang, \\luoyuanchang1, weidaimeng yanghao30\}@huawei.com}


\begin{document}

\maketitle
\def\thefootnote{*}\footnotetext{These authors contributed equally to this work}\def\thefootnote{\arabic{footnote}}

\begin{abstract}
    
    
    
    Abstractive Speech Summarization (SSum) aims to generate human-like text summaries from spoken content. It encounters difficulties in handling long speech input and capturing the intricate cross-modal mapping between long speech inputs and short text summaries. Research on large language models (LLMs) and multimodal information fusion has provided new insights for addressing these challenges. In this paper, we propose an end-to-end SSum model that utilizes Q-Former as a connector for the audio-text modality and employs LLMs to generate text summaries directly from speech features. We adopt a multi-stage training approach that includes LLM based ASR and Text Summarization (TSum) tasks as auxiliary tasks. ASR tasks are used to align feature spaces and enhance the LLM's ability to handle longer speech. Then, we utilize a curriculum learning strategy to facilitate the model's transition from TSum to SSum. Finally, our model achieves competitive performance on the How-2 dataset.
\end{abstract}

\section{Introduction}
Abstractive Speech Summarization (SSum) \cite{murray2010generating,sharma2021index,finley2018automated} aims to directly generate human-friendly textual summaries from relatively long speech inputs. Compared to Text Summarization task \cite{neto2002automatic}, its core challenges are: (a) the long speech sequences pose a computational complexity bottleneck; (b) the non-monotonic and complex mapping between long speech inputs and short text summaries; (c) the modality gap between audio inputs and text outputs. Previous methods can be categorized into two types: cascaded models \cite{palaskar2019multimodal,shon2022slue,palaskar2021multimodal} of Automatic Speech Recognition (ASR) and Text Summarization (TSum), or end-to-end models \cite{sharma2022end,kano2023speech,matsuura2023leveraging,sharma2023bass}. Recent research has shown that end-to-end models can outperform cascaded systems \cite{sharma2022end,matsuura2023leveraging}, as they can extract para-linguistic information from speech and address the error propagation issue in cascaded systems. However, in order to encode long audio directly, end-to-end models often need to truncate audio, or utilize restricted attention \cite{sharma2022end, beltagy2020longformer} or alternatives like F-Net \cite{kano2023speech}, which limit further model improvements. 

Recently, the rapid progress of large language models \cite{achiam2023gpt,touvron2023llama,brown2020language} has drawn interest from multiple research areas due to their capacity for handling extremely long inputs and excellent performance in NLP tasks like question answering, reasoning, and summarization. Speech processing is adopting the latest advancements from LLMs, including tasks such as ASR \cite{yu2023connecting}, GPT-style speech language models \cite{zhang2023speechgpt}, and a range of other applications \cite{rubenstein2023audiopalm}, all leveraging the benefits of using LLMs in this field. To integrate speech features into LLMs, a connector is typically required, where Querying Transformer (Q-Former) \cite{li2023blip} has been proven to be a relatively efficient cross-modal information extraction method \cite{yu2023connecting}. It can convert variable-length input sequences into fixed-length output query representations. We believe that by integrating Q-Former for cross-modal encoding between speech and text, and leveraging LLMs to manage tasks like processing long input speech and creating concise summaries, we can further improve the model's performance in end-to-end speech summarization. 

\begin{figure}[t]
  \centering
  \includegraphics[width=\linewidth]{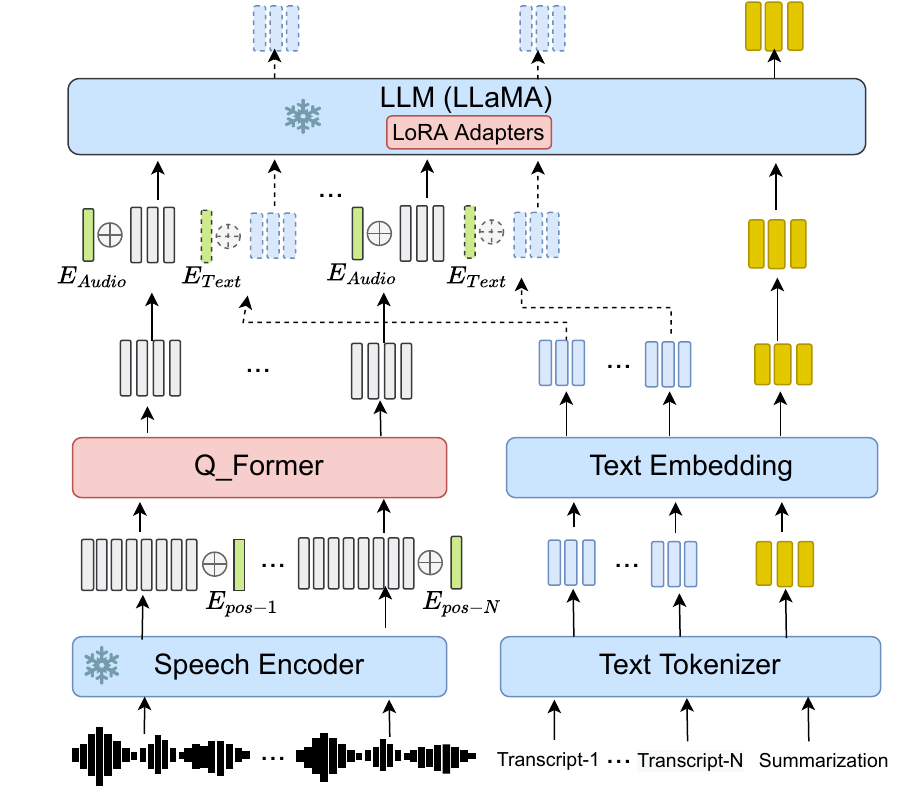}
  \caption{Overview of the proposed model. A speech encoder and Q-Former are used to extract speech features. LLM accepts speech prompts and generates text summaries directly. Text transcripts are used as auxiliary information during the training.}
  \label{fig:model_overview}
\end{figure}

As traditional transformer-based speech encoders find it challenging to handle longer speech, it is intuitive to segment the speech for encoding and then connect the feature segments to build the final representation. In this paper, we attempt to integrate long speech inputs into LLMs using segment level Q-Former and train a LLM based end-to-end speech summarization model through efficient parameter fine-tuning method. In detail, we utilize a speech encoder and Q-Former to extract speech features for individual segments of long speech. Then, we combine the speech features from all segments and feed them into LLM. Finally, LLM employs these speech features as prompts to generate the ultimate text summaries in an autoregressive manner. The proposed model's overview can be found in Figure \ref{fig:model_overview}. However, our model still faces the following challenges:
\begin{enumerate}
\item The output of Q-Former needs to be aligned with the input of LLM so that LLM can recognize the speech features. 

\item The speech segmentation strategy may hamper the model's capability to handle the context of long speech, as there is no interaction between segments during encoding.

\item Compared to text summarization tasks, speech summarization still faces the modality gap between speech and text.
\end{enumerate}

To tackle these challenges, our model initial aligns the Q-Former output with the LLM input effectively via a sentence-level ASR task. Then, we improve the model's ability to handle longer speech by incorporating a Document-level ASR task. Finally, to further bridge the gap between modalities, we conduct joint training on two tasks, TSum and SSum using a curriculum learning approach \cite{bengio2009curriculum}.

We validated our proposed model on the widely used How2 \cite{sanabria2018how2} dataset. Our experiments demonstrate that our multi-stage training strategy effectively prepares LLMs for end-to-end speech summarization tasks by leveraging ASR and TSum tasks. The final model's performance exceeds that of the cascaded models and is comparable to the strong baselines of traditional end-to-end models based on the BERTScore metric.

\section{Related Work}
Speech Summarization \cite{murray2010generating,sharma2021index,finley2018automated} can be tackled using either cascaded or end-to-end methods, with each approach having its own strengths and weaknesses based on the particular application scenario. Initially, cascaded systems \cite{palaskar2019multimodal,shon2022slue,palaskar2021multimodal}, leveraging pre-trained ASR and TSum models, can be individually enhanced with domain-specific \cite{matsuura2023transfer} data before cascading them together to generate the final text summary. Studies have proven that cascaded systems can achieve competitive performance but also face challenges such as error propagation, longer inference delays, and the inability to fully utilize audio information. On the other hand, end-to-end systems \cite{sharma2022end,kano2023speech,matsuura2023leveraging,sharma2023bass} can abstract text summaries from speech features and have been shown to outperform cascaded models on some datasets. However, when dealing with long speech recordings, input truncation or non-standard self-attention modules are essential \cite{sharma2022end, beltagy2020longformer,kano2023speech}. Additionally, \cite{jung2024augsumm} explored using large models to construct more enriched summary labels to enhance model performance. As far as we know, there is currently no direct effort to convert LLMs into end-to-end Speech Summarization models. 

\section{Methodologies}

In Figure \ref{fig:model_overview}, we present an overview of the proposed model, which comprises three main components: a speech encoder, a Q-Former module, and a LLM. The model training is divided into three stages to allow the model to bridge the modality gap and achieve better performance.

\subsection{Speech feature extraction}

Initially, we need a speech encoder denote as S-Encoder, which can be pretrained or trained from scratch, for extracting speech features from the raw waveform. For clarity, let's define some key notations: $X \in R^{n_x \times d_x}$ represents the speech features extracted from the S-Encoder, where $n$ and $d$ are the numbers of vectors and hidden dimensions respectively.

Q-Former is responsible for further compressing $X$ into a fixed-length representation $Q \in R^{n_q \times d_q}$, serving as the final input feature for the LLM. Notably, we included a weighted sum module in the model to help Q-Former extract a wide range of speech features, enabling the model to leverage useful signals in the speech aside from text. 

For longer audio inputs, we segment the speech into segments and introduce segment-level position embedding $E_{pos}$ to $X$ so that Q-Former can learn the positional information of different segments.  So the the final speech feature $F_{speech}$ can be calculated as follows:
\begin{equation}
F_{speech} = [Q\text{-}Former(S\text{-}Encoder(x_i) \oplus E_{pos})]_{i=1}^{N}
\end{equation}

\subsection{LLM for end-to-end Speech Summarization}
We choose LLaMA2-7B \cite{touvron2023llama} as the base LLM and employ the parameter-efficient Low-rank Adaptation (LoRA) \cite{hu2021lora} to fine-tune the model, while keeping other LLM parameters frozen. The speech features $F_{speech}$ are used as prompt tokens to guide the model to generate text summaries $T_{sum}$ directly in an autoregressive manner. The transcript text ($T_{trans}$) corresponding to the speech is utilized as auxiliary information during the training process. Notably, we introduce embeddings ($E_{audio}$ and $E_{text}$) to differentiate the modality information of different input features, thereby helping LLM bridge the modality gap. Therefore, the final loss we optimize is as follows: 
\begin{equation}
\mathcal L_{LM} = -\sum_{i=1}^{T_{sum}} \log P(y_i|y_{<i}, F_{speech} \oplus E_{audio};\theta_{LoRA})
\end{equation}

\subsection{Training Strategy}
\begin{figure}[t]
  \centering
  \includegraphics[width=\linewidth]{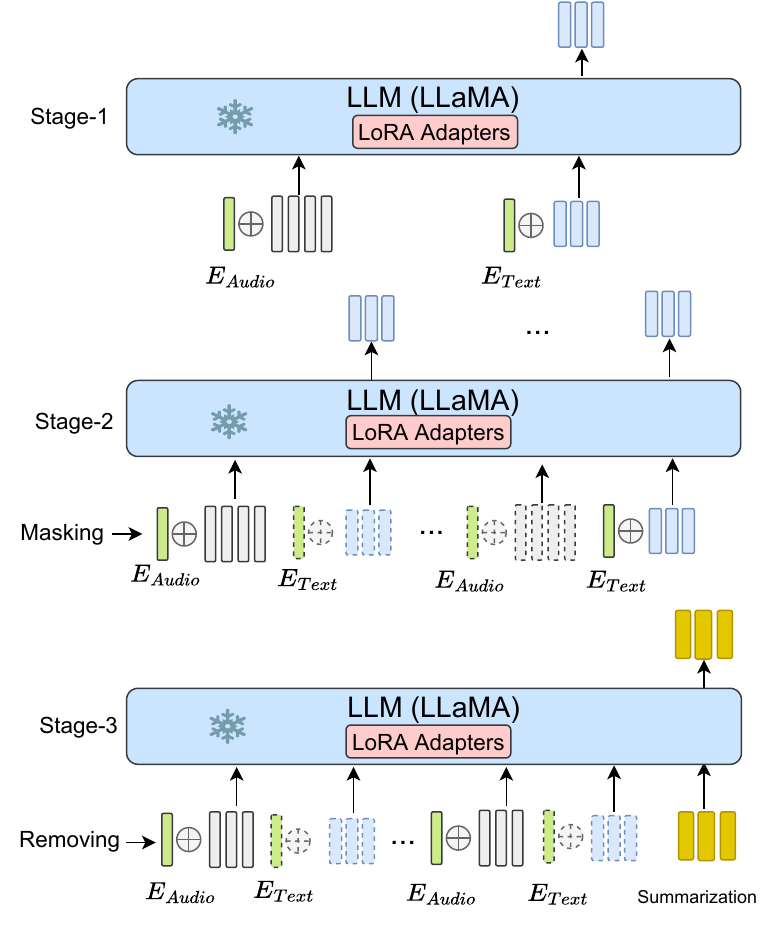}
  \caption{The training pipline of our model.}
  \label{fig:training}
\end{figure}

The training is divided into three stages, and the schematic diagrams of the model inputs and outputs for each stage are shown in Figure \ref{fig:training}:

\begin{enumerate}
\item First, we train a sentence-level ASR model where Q-Former extracts speech feature tokens and LLM uses them as prompts to generate corresponding transcriptions. Each segment of the speech is optimized separately without interaction between segments and without the need for segment positional embeddings.

\item Next, we then flatten the speech features ($F_{speech}$) and transcription features ($T_{trans}$) from various segments of a long speech recording to train a document-level ASR task. This approach promotes contextual connections among segments, improving the model's capability to understand extended speech contexts. Additionally, randomly masking speech or transcription features within a segment helps align speech and text representations.

\item Finally, we optimize the ultimate end-to-end Speech Summarization task. Training directly on SSum still faces the modal gap issue compared to the TSum task. Therefore, we employ the concept of curriculum learning (CL) to gradually transition the model from the TSum task to the SSum task. At the beginning, the model utilizes all speech and text features to complete the summarization task, and the model's input aligns with Stage 2. Subsequently, we progressively remove the transcribe text features until only speech features remain.
\end{enumerate}

\section{Experimental Setup}
In this section, we will discuss the details of our experiments, including the dataset, model configurations, evaluation metrics, and so on.

\subsection{Dataset}
\begin{table}[th]
  \caption{Statistics of the How-2 2,000h Dataset used for model training and evaluation. The input length N (in frames), and output length L (in tokens) are shown.}
  \label{tab:how2}
  \centering
  \begin{tabular}{cllcccc }
    \toprule
    {\textbf{Set}} & {\textbf{Max N}} & {\textbf{Mean N}} & {\textbf{Mean L}} & {\textbf{Max L}} \\
    \midrule
    Train & 145,082 & 9,806.6 & 60.5 & 173 \\
    Test & 39,537 & 9,866.6 & 60.3 & 152\\
    \bottomrule
  \end{tabular}
\end{table}

The How-2 Dataset, as outlined in \cite{sanabria2018how2}, contains 2,000 hours of instructional videos accompanied by text transcripts, video content, speech, translations, and summaries. Abstractive summaries are generated based on speech for an end-to-end approach. Table \ref{tab:how2} presents the statistics for the training and testing partitions of the How2 dataset. The model features and reference summaries can be found here \footnote{https://github.com/srvk/how2-dataset}. At the same time, we merged the original speech segments in the dataset and kept the length of each individual speech segment to around 30 seconds to enhance encoding efficiency. 

\subsection{Model and Training configurations}
The core components of our model are as follows: 

\textbf{Speech Encoder: } We begin by training a standard ASR model using an attention-based sequence model, comprising a 12-layer conformer encoder and a 6-layer transformer decoder.  The training loss is a hybrid CTC/Attention, with a CTC weight of 0.3. The model utilizes hidden and feedforward dimensions of 768 and 3072, respectively. We use the encoder of the ASR model as the speech encoder and keep it frozen during subsequent training. 

\textbf{Q-Former:} Our Q-Former module inherits the settings from \cite{li2023blip}, starting with a pretrained $BERT_{base}$ \cite{devlin2018bert} and keep updating during training. There are 150 trainable queries for each speech segment. Then, we concatenate the outputs of Q-Former to align with the input feature dimensions of LLM. Finally, for approximately 30 seconds of speech, the number of speech feature tokens is also 30. 

\textbf{LoRA adapter for LLM:} We use the LoRA approach to adapt the key, query, value and output layers of the self-attention mechanism leaving other part of LLaMA2-7B model unchanged . Unless specified otherwise, default LoRA hyperparameters are set to a rank of $R$ = 8 and $a$ = 16.

\textbf{Baseline systems:} We compare two baseline systems: one uses ground truth (GT) transcripts, while the other incorporates ASR transcripts along with LLM for summarization generation.

During training, the Huggingface transformers library \footnote{https://huggingface.co/docs/transformers/en/index} and 8 GPUs are used in all of our experiments. When training the Speech encoder, adam optimizer is used with a peak learning rate of 0.002 in 100k training steps and the batch size is 128.  

For the training of end-to-end models, we still use Adam optimizer with a learning rate of 2e-4, warmup steps of 8k, and a total training step of around 100K. Additionally, an early stop strategy is employed to prevent overfitting. For different stages of training, we adjust the parameters for gradient accumulation to maintain a batch size of 128.

In the second training stage, we set the random masking probability to 0.2. When training in stage 3 with curriculum learning, we dedicate 20\% of the training steps to jointly optimizing TSum and SSum, 50\% to curriculum learning, and the final 30\% to training the SSum task exclusively.

\subsection{Metrics}
We evaluate our models with ROUGE \cite{lin2004rouge}, METEOR \cite{banerjee2005meteor}, and BERTScore \cite{zhang2019bertscore}, which are the most common automatic metrics for evaluating summarization models.

\section{Result and  Analysis}
\subsection{Main Result}

\begin{table}[th]
  \caption{The main results of our experiment. The baseline models used for comparison include: LLM-based cascading models and the end-to-end models from prior work. GT and ASR mean ground truth and ASR transcript, respectively. }
  \label{tab:example}
  \centering
  \begin{tabular}{llcccc }
    \toprule
    {\textbf{Models}} & {\textbf{ROUGE-1, 2, L}} & {\textbf{MTR}} & {\textbf{BTS}} \\
    \midrule
    \textbf{Cascade} &&& \\
    LLaMA2-7B + GT  & 64.3, 49.1, 60.4 & 33.7 & 93.93 \\
    LLaMA2-7B + ASR & 61.3, 46.6, 58.6 & 32.1 & 91.76 \\
    \hline
    \textbf{Previous E2E} &&& \\
    Longformer \cite{sharma2022end} & 60.7, 44.9, 56.1 & 29.3 & 91.53 \\
    FNet \cite{kano2023speech} & 61.9, 42.3, 58.8 & 29.0 & - \\
    Standard AT \cite{matsuura2023leveraging} & 65.3, 51.4, 62.1 & 32.5 & 93.00 \\
    \quad  + TTS & 68.4, 54.1, 65.0 & 34.9 & 93.80 \\
    BASS \cite{sharma2023bass} & 64.0, 49.0, 60.2 & 32.2 & 92.51 \\ 
    Pre-trained LM \cite{matsuura2023transfer} & 67.0, 52.1, 63.2 & 34.4 & 93.98 \\
    \hline
    \textbf{ours E2E}  &&& \\
     QF + LLaMA2-7B  & 63.8, 48.4, 59.7 & 33.4 & \textbf{93.85}\\
     \quad w/o stage-2 & 63.1, 47.6, 59.2 & 32.7 & 93.47\\
     \quad w/o CL+TSum & 62.7, 47.2, 58.2 & 32.4 & 92.87\\
     QF + LLaMA2-13B  & 64.1, 48.9, 59.4 & 33.5 & \textbf{93.88}\\ 
    \bottomrule
  \end{tabular}
  
\end{table}

Table 2 summarizes some of our experimental results, including baseline models for cascade method and typical end-to-end models from previous works. Our end-to-end model exceeds the baseline cascade system using ASR transcripts and LLaMA2-7B in various evaluation metrics, and even on par with systems using ground truth transcripts and LLaMA2-7B in the BERTScore metric. This demonstrates that our model has successfully mitigated the error propagation effects caused by ASR systems, and has successfully bridged the modality gap. 

When compared to some highly optimized strong end-to-end models in the past (utilizing TTS data augmentation, text summarization data), although our model shows a certain gap in ROUGE and METEOR metrics, it can essentially match them in terms of BERTScore. This also demonstrates the advantage of LLM in high-level semantic summarization capability. Relevant data augmentation and further optimization work are left for future exploration.

The ablation experiments also demonstrate that the training in Stage 2 and the curriculum learning with the TSum task used in Stage 3 contribute to the final results of the model, with the latter being the more crucial factor. 

In the end, we attempted to use a larger LLaMA2-13B model to improve the summarization performance. However, we only observed an improvement in the ROUGE-1,2 metric, while the other metrics remained consistent with the LLaMA2-7B model. This may indicate that the 7B model is already sufficient to address the current task, or that larger models may have other training-related problems, which will also be explored in the future. 
\subsection{Analysis}
To further analyze our models, we conducted the following additional experiments: 

\textbf{Alignment for speech features:} In training stage 1, we attempted to align the output of Q-Former with the input of LLM through an ASR task, allowing us to obtain an LLM based ASR model (LLM-ASR). The performance of this model can to some extent measure the effectiveness of feature alignment. We summarized some WER results of different models in Table \ref{tab:asr_result}. The Base-ASR comes from our own trained basic ASR model, while Baseline \cite{sharma2022end} is from previous work. The results indicate that we have obtained a competitive ASR model, with a decrease of 0.3 compared to the base ASR, showing that the features extracted by Q-Former can be recognized by the LLM.

\begin{table}[th]
  \caption{Sentence level Word Error Rate (WER) (\%) for Test sets of the 2000h How-2 Corpus.}
  \label{tab:asr_result}
  \centering
  \begin{tabular}{lllccc }
    \toprule
    {\textbf{Model}} & {\textbf{Encoder}} & {\textbf{Decoder}} & {\textbf{WER (\%)}} \\
    \midrule
    Baseline \cite{sharma2022end} & Conformer  & Transformer & 9.1 \\
    Base-ASR & Conformer & Transformer & 8.8 \\
    \hline
    LLM-ASR & Conformer & LLaMA2-7B & 8.5 \\
    \bottomrule
  \end{tabular}
  
\end{table}

\textbf{Long speech context learning:} In training stage 2, we attempted to leverage document-level ASR tasks to enhance the modeling capability of long speech recordings, enabling us to obtain an LLM based document level ASR model (LLM-DOC-ASR). If we successfully achieve our goal, the recognition capability of long speech recordings by the model will also be improved. To validate our hypothesis, we compared the WER of the two ASR models obtained from training stage 1 and training stage 2, as well as the Perplexity (PPL) of the transcribed text at the document level. The results in Table \ref{tab:long_audio} show that both WER and PPL have been optimized, proving that LLM can more effectively handle long speech inputs after training in Stage 2.

\begin{table}[th]
  \caption{The WER and document-level PPL results of transcription texts from different models on the how-2 test set. }
  \label{tab:long_audio}
  \centering
  \begin{tabular}{llcc }
    \toprule
    {\textbf{Model}} & {\textbf{WER (\%)}} & {\textbf{PPL}}\\
    \midrule
    Base ASR  & 8.8 & 6.4 \\
    LLM-ASR  & 8.5 & 6.2 \\
    LLM-DOC-ASR & 8.2 & 5.8 \\
    \bottomrule
  \end{tabular}
\end{table}

\begin{figure}[t]
  \centering
  \includegraphics[width=\linewidth]{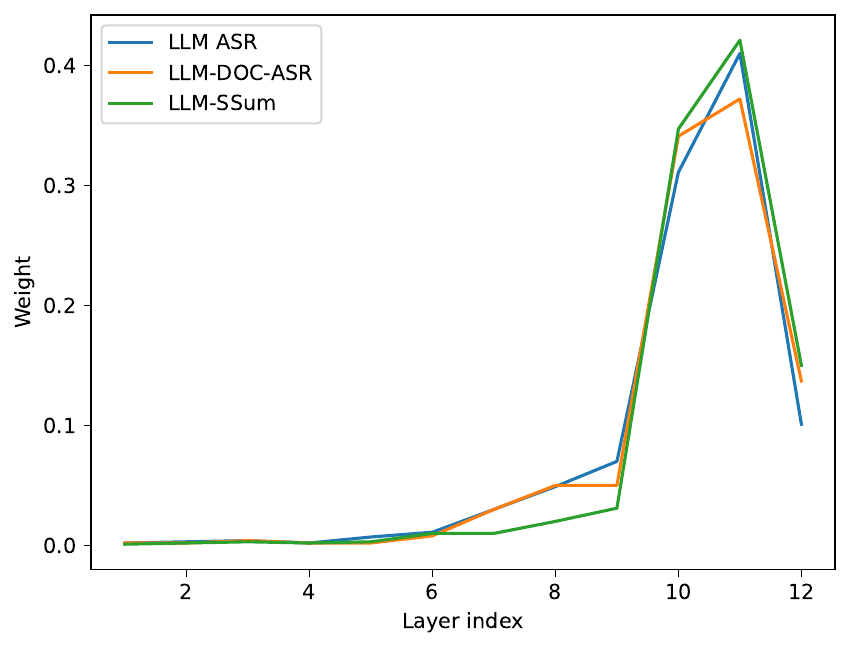}
  \caption{The weight distribution in the weighted sum module of different models obtained from different training tasks.}
  \label{fig:weight_sum}
\end{figure}

\textbf{Weight distribution for tasks:}
In order to explore how our model utilizes different levels of speech features in various training tasks, we analyzed the weight distribution in the weighted sum module for the speech encoder. The weight distribution is plotted in Figure \ref{fig:weight_sum}. Overall, the model tends to use high-level features more, whether it is for the early ASR task or the later Summarization task. This indicates that modality conversion tasks cannot benefit from low-level features. Nevertheless, we can still observe that, compared to the ASR task, the SSum task prefers higher-level abstract semantic information, with relatively higher weights for the last three layers. 

\section{Conclusion}
In this work, we attempt to combine the cross-modal feature extractor Q-Former with the LLMs to solve the end-to-end speech summarization task. To achieve our goal, we segment long speech and extract speech features using Q-Former, then guide LLMs to generate summaries directly. Afterwards, we use ASR and the TSum task as auxiliary tasks and divide the training into multiple stages to overcome challenges faced by the model such as feature space alignment, understanding long speech, and cross-modal mapping. Finally, we validate our model on the how2 dataset.

\bibliographystyle{IEEEtran}
\bibliography{mybib}

\end{document}